%% file: ms.tex
\pdfoutput=1 

\documentclass[10pt,journal,compsoc]{IEEEtran}
\ifCLASSOPTIONcompsoc
\usepackage[nocompress]{cite}
\else
\usepackage{cite}
\fi
\ifCLASSINFOpdf
\else
\fi
\usepackage{tabularx}
\usepackage{arydshln}
\usepackage{paralist}
\usepackage{float}
\usepackage{multirow}
\usepackage{perpage} 
\usepackage{times}
\usepackage{epsfig}
\usepackage{array,booktabs}
\usepackage{mathtools}
\DeclarePairedDelimiter{\ceil}{\lceil}{\rceil}
\usepackage{wrapfig}

\usepackage{graphicx}

\usepackage{epstopdf}
\usepackage{xfrac}
\usepackage{xr}
\usepackage{tikz}
\usepackage{pgfplots}
\pgfplotsset{compat=1.11}
\usepackage{tkz-euclide}
\usetkzobj{all}
\usetikzlibrary{intersections,decorations.pathreplacing,positioning,calc, patterns, quotes,angles, arrows,shapes}

\usepackage{algorithm}
\usepackage{algorithmic}

\newlength\myindent
\usepackage{amsmath}

\usepackage{amssymb}

\usepackage{amsthm}

\usepackage{bm}

\newtheorem{theorem}{Theorem}

\theoremstyle{definition} 

\usepackage[caption=false,font=footnotesize]{subfig}
\usepackage{booktabs}

\usepackage{float}

\usepackage{placeins}

\usepackage{verbatim}					

\usepackage{acronym}

\usepackage{color}

\usepackage{url}

\usepackage{cite}

\DeclareMathOperator{\Tracetmp}{trace}
\newcommand{\trace}[1]{\Tracetmp \left\{ #1 \right\} }

\definecolor{gray}{rgb}{0.5,0.5,0.5}	
\definecolor{red}{rgb}{1.0,0.0,0.0}	  
\definecolor{green}{rgb}{0.0,1.0,0.0}	  
\definecolor{blue}{rgb}{0.0,0.0,1.0}	  

\newcommand{\eg}{\textit{e.g.}}

\newacro{SVM}{Support Vector Machine}
\newacro{MM}{Majorization-Minimization}
\newacro{*SHL}{*Supervised Hash Learning}
\newacro{BCD}{Block Coordinate Descent}
\newacro{psd}{positive semi-definite}

\newacro{COP}[CoP]{Coherence Pursuit}
\newacro{KCOP}[GCP]{Global Conformity Pursuit}
\newacro{PCA}{Principal Component Analysis}
\newacro{KPCA}[Kernel PCA]{Kernel PCA}
\newacro{LDA}{Linear Discriminant Analysis}
\newacro{RRQR}{Rank-Revealing QR}
\newacro{L1KPCA}[$L_1$-KPCA] {$L_1$-norm Kernel PCA}

\newacro{FDR}{Outlier Detection in the Frame of Dimensionality Reduction}
\newacro{RML}{Outlier Detection for Robust Manifold Learning}
\newacro{SVD}{Singular Value Decomposition}
\newacro{pd}{positive-definite}
\newacro{ML}{Manifold Learning}
\newacro{MC}[GC]{Global Conformity}

\newacro{name}[MOSAIC]{\textbf{M}anif\textbf{O}ld \textbf{SA}mpling and outlier \textbf{I}dentification based on representativeness \textbf{C}ertification}
\newacro{ADMM}{Alternating Direction Method of Multipliers}
\newacro{OP}[\textit{OP}]{\textit{Outlier Probability}}
\newacro{RS}[\textit{RS}]{\textit{Representativeness Score}}
\newacro{SM}[\textit{SM}]{\textit{Sparsity Measure}}

\newacro{DS3}{Dissimilarity-based Sparse Subset Selection}
\newacro{SMRS}{Sparse Modeling Representative Selection}
\newacro{SSM}{Spectral Sampling of Manifolds}
\newacro{Kmed}[Kmedoids]{Kmedoids}
\newacro{M-kmeans}{Manifold K-means}

\newacro{MDS}{Multi-Dimensional Scaling}
\newacro{MMDS}[Metric MDS]{Metric Multi-Dimensional Scaling}
\newacro{LS-MDS}[LS-MDS]{Least Squares Multi-Dimensional Scaling}
\newacro{SM}{Sammon Mapping}
\newacro{GD}{Gradient Descent}
\newacro{IM}{Iterative Majorization}
\newacro{DML}{Distance Metric Learning}
\newacro{RKHS}{Reproducing Kernel Hilbert Space}
\newacro{K-LS-MDS}[Kernel-based LS-MDS]{Kernel-based Least Squares Multi-Dimensional Scaling}
\newacro{SIMDS}[SI-MDS]{Sparse Inductive Multi-Dimensional Scaling}
\newacro{MKL}{Multiple Kernel Learning}

\newacro{CG}{Conjugate Gradient}
\newacro{MMO}{Multi-level Modularity Optimization}
\newacro{APG}{Accelerated Proximal Gradient}

\newcommand{\eref}[1]{Equation~(\ref{#1})}
\newcommand{\fref}[1]{Fig.~\ref{#1}}
\newcommand{\tref}[1]{Table~\ref{#1}}
\newcommand{\sref}[1]{Section~\ref{#1}}

\newcommand{\aref}[1]{Algorithm~\ref{#1}}

\newcommand{\thmref}[1]{Theorem~\ref{#1}}

\begin{document}
	%
	\title{A Multi-criteria Approach for Fast and Outlier-aware Representative Selection from Manifolds}
	%
	%
	%
	\author{Mahlagha~Sedghi,~\IEEEmembership{Student~Member,~IEEE}, George~Atia,~\IEEEmembership{Senior~Member,~IEEE}, and Michael~Georgiopoulos,~\IEEEmembership{Senior Member,~IEEE}
		\IEEEcompsocitemizethanks{\IEEEcompsocthanksitem The authors are with the Department of Electrical and Computer Engineering, University of Central Florida, Orlando, FL, 32816, USA. \protect\\
			E-mails: mahlagha.s@knights.ucf.edu, George.Atia@ucf.edu, and michaelg@ucf.edu
	}%
	}

	\IEEEtitleabstractindextext{%
		\begin{abstract}
			\input{Abstract}
		\end{abstract}
		
		\begin{IEEEkeywords}
		Representative Sampling, Manifold Learning, Outlier-aware Subset Selection, Outlier Detection, Kernel Methods.
	\end{IEEEkeywords}}

	\maketitle

	\IEEEdisplaynontitleabstractindextext

	%
	\IEEEpeerreviewmaketitle

	\IEEEraisesectionheading{\section{Introduction}\label{sec:introduction}}

	%
	%
	%
	%
\input{Introduction}
\label{sec:Introduction}
\section{Proposed Method}
\label{sec:Method}
\input{Method}
\section{Algorithm}
\label{sec:Algorithm}
\input{Algorithm}
\section{Theoretical Analysis}
\label{sec:Analysis}
\input{Analysis}
\section{Experiments}
\label{sec:Experiments}
\input{Experiments}
\section{Conclusion}
\label{sec:Conclusion}
\input{Conclusion}

\bibliographystyle{ieeetr}
\bibliography{Reference}

\end{document}

%% file: Abstract.tex
The problem of representative selection amounts to sampling few informative exemplars from large datasets.
This paper presents MOSAIC, a novel representative selection approach from high-dimensional data that may exhibit non-linear structures. Resting upon a novel quadratic formulation, Our method advances a multi-criteria selection approach that maximizes the global representation power of the sampled subset, ensures diversity, and rejects disruptive information by effectively detecting outliers. Through theoretical analyses we characterize the obtained sketch and reveal that the sampled representatives maximize a well-defined notion of data coverage in a transformed space. In addition, we present a highly scalable randomized implementation of the proposed algorithm shown to bring about substantial speedups. MOSAIC's superiority in achieving the desired characteristics of a representative subset all at once while exhibiting remarkable robustness to various outlier types is demonstrated via extensive experiments conducted on both real and synthetic data with comparisons to state-of-the-art algorithms.

%% file: Introduction.tex
Representative selection entails selecting few data points out of the original collection that are simultaneously representative and concise thereby reducing the problem size.
Besides considerable resource conservations offered by subset selection approaches, summaries afforded by informative representatives can also facilitate insightful interpretations of big data and complex systems. Moreover, working with a small subset alleviates the need for burdensome tasks of data annotation by human resources.
Data sketching has bearing on many applications, including natural language processing, recommender systems, computer vision, medicine, and marketing, and social networks \cite{Wang2015, Li2019, Hartline2008, Liu2015a, Song2018}. 

Brute force search over all possible subsets of a dataset, however, is NP-hard, hence intractable for big data. The problem is more challenging in the presence of outliers that prevail much of today's data, since the selection mechanism has to effectively reject outliers while selecting the most informative data exemplars. While there have been noteworthy efforts to develop efficient and robust techniques to tackle this problem, 
this paper is motivated by several important limitations of existing approaches, such as inability to handle non-linear data, sensitivity to outliers, lack of interpretability, and choosing single-criterion samples such as only diverse or only informative subsets.

One recurring issue emerges from the linear models which often assume a low-rank structure for the data, making them inapplicable for non-linear settings. These methods mainly attempt to span the column space of the original dataset, and have been studied widely in the literature such as in \cite{Gu1996,Mahoney2011,Elhamifar2012a}. To improve the scalability, some have attempted to yield approximation solutions such as \cite{Boutsidis2009,Frieze2004}.
In presence of non-linear patterns, employing low-dimensional manifold structures has shown promise in a plethora of machine learning and signal processing tasks such as in dynamical systems, network information management and clustering \cite{Guo2008,Gong2012,Wu2015,Sedghi2017,Feng2018}. 
In order to eliminate the restriction on the allowable data structures, few methods tackle the sampling problem with a \ac{ML} approach\cite{Oeztireli2010, Shroff2011,Du2013,Ye2015}.
 These methods either adopt graph-based distances as approximate measures of geodesic distances, replacing the Euclidean distance in the conventional linear approaches, or approximate local neighborhood sets of manifolds by linear subspaces and then apply the linear models to those sets. Consequently, they inherit the deficiency of the original methods, and incorporate local information, which diminishes their ability to capture a global view of the underlying structure. We propose an approach which enables representative sampling from data adhering to non-linear manifold structures by a different means. 
We obtain a concise encoding of the global manifold structure that underpins a multi-criteria paradigm through a novel convex formulation.

Another limitation is the inability of many of the existing methods to successfully reject outliers in the selected subsets of representatives. Although some incorporate outlier identification strategies, it remains a challenge to yield sets devoid of outliers.  
This issue becomes more prominent in high-dimensional settings where the outliers are more easily concealed 
due to the so-called curse of dimensionality. 
Robustness to outliers is naturally enabled in our approach through a sparsity measure of the obtained encoding, which gauges the conformity of a data point to the entire data collection.

Also, an important concern is the lack of interpretability in the sense of being able to explain the choice of specific data points as data representatives by many of the existing algorithms (e.g., algebraic approaches) -- this has become more momentous in recent years considering the attention accorded to interpretable/explainable machine learning. Here, we provide an analysis rooted in geometric functional analysis, which gives insight into the characteristics of the representative set.

Lastly, many techniques approach the problem taking only a single criterion into account which limits their efficacy in practice. Clustering-based methods, for instance \cite{Elhamifar2014, Frey2006, Kaufman1987, Charikar2002}, build upon group-forming patterns and disregard the underlying structure of the data, whereas the diversity-driven ones focus solely on maximal diversity of the subset rather than its representativeness \cite{Kulesza2011, Farahat2015}. As a result, the selected subsets based on such algorithms are not always descriptive summaries of the whole set.

\subsection{Contributions}
In this paper, we make four main contributions. First, motivated by the aforementioned limitations, we develop a novel method for robust sampling of representatives from high-dimensional data, which effectively captures non-linear behavior of data lying on manifold structures. Our proposed method, dubbed \ac{name}, builds upon an obtained matrix which contains representation power of each sample with regards to the others.  
This encoding is obtained from the solution of a newly formulated quadratic program, which not only yields a global optimum independent of the initialization, but also is equipped with a computationally efficient algorithm.

As our second contribution, we introduce a three-pronged approach derived from this concise encoding, which considers a comprehensive set of features essential for a representative subset, namely \emph{representativeness, conciseness, and robustness}. 
The first criterion ensures that the sampled subset is sufficiently informative to be able to represent the whole dataset with as much fidelity as possible, while the second one ensures diverse information. These two conditions yield a minimalistic set of descriptive samples, maximizing the amount of information per chosen sample. Lastly, to avoid misinformation, the sampled representatives should not contain outlying data points, which is satisfied by our third criterion. 
Our third contribution lies in deriving a theoretical characterization of the representatives identified by \ac{name}. In a nutshell, our selected representatives are fully characterized through a feature space, and are shown to maximize a well-defined notion of coverage for the whole data set. 
The analysis leverages a profound connection to the theory of Reproducing Kernel Hilbert Spaces established in the supplementay material, which affords an insightful re-interpretation of our methodology.

As our fourth contribution, we develop a scalable implementation of the algorithm proposed, which brings about substantial speedup enabling the processing of large datasets. Representative selection is only applied to a sketch of the data that is adaptively augmented through an iterative refinement procedure to successively incorporate underrepresented data points (See \sref{sec:scalable}).  

We remark that the proposed method offers a stand-alone outlier detection technique for high-dimensional data. Indeed, the procedure effectively yields a two-way decomposition of the data, generalizing the low-rank plus sparse matrix decomposition principle \cite{Cand`es2011} to a comprehensive paradigm capable of simultaneously dealing with linear/non-linear settings and non-sparse outlier models. Although it is not the primary focus of this paper, we include additional experimental results in \sref{sec:outlier_exp}, which show that \ac{name} remarkably outperforms other state-of-the-art outlier detection methods.

To the best of our knowledge, \ac{name} is the first manifold learning approach that addresses all the aforementioned shortcomings at once. It demonstrates considerable superiority compared to state-of-the-art methods through a variety of experiments including classification and clustering tasks, tolerating random and structured outliers, and algorithm running time. 
\subsection{Notation}
\label{sec:Notation}

  Let $\mathbb{N}_k \triangleq \left\{ 1, \ldots, k \right\}$ for $k \in \mathbb{N}$.
  Column vectors and matrices are denoted in boldface lower-case and upper-case letters, respectively. Also, let $\mathbf{1}$ and $\mathbf{I}_n$  denote the all-ones vector of proper length and the identity matrix of size $n$, respectively. 
  For a vector $\mathbf{a}$, $\left\| \mathbf{a} \right\|_p$ stands for its $\ell _p$-norm, and $\mathbf{a}(i)$ denotes its $i$\textsuperscript{th} element. For a matrix $\mathbf{A}$, $\mathbf{a}_i, a_{ij}$ denote its $i$\textsuperscript{th} column and $(i,j)$\textsuperscript{th} element, respectively. Also, $\left\| \mathbf{A} \right\|_F$ denotes its Frobenius norm, and $\left\| \mathbf{A} \right\|_{1,p}$ stands for its group Lasso norm defined as $\left\| \mathbf{A} \right\|_{1,p} = \sum_i \left\| \mathbf{a}_i \right\|_p$.
 

%% file: Method.tex
In this section, we present the \ac{name} method, which affords a simple yet powerful approach for robust sampling of representatives from non-linear manifolds. 
Inspired by many real-world scenarios, we consider the case where the data does not necessarily behave in a linear fashion, rather, it can also belong to some low-dimensional non-linear manifolds, and contain outlying data points. Hence $n$ data points are arranged as a data matrix $\mathbf{D} \in \mathbb{R}^{m \times n}= [\mathbf{A} \ \mathbf{B}] \ \mathbf{T}$, where $\mathbf{A}\in \mathbb{R}^{m \times n_1}$ represents the inliers lying on manifolds, $\mathbf{B} \in \mathbb{R}^{m \times n_2}$ stands for the collection of outlying points, and $\mathbf{T}$ is an arbitrary permutation matrix. 

We will first establish our informative encoding of the data relations in \S \ref{sec:encoding}, using which the method is capable of satisfying the three aforementioned criteria in \S \ref{sec:representative} - \S \ref{sec:outliers}. Later, a profound connection of our formulation with kernel methods is established in the supplementary material, which gives a different perspective on our methodology, and facilitates interesting theoretical analysis in \S \ref{sec:Analysis}. 
\subsection{Representation Power Encoding}
\label{sec:encoding}
Consider we are given a matrix of pairwise similarities $\mathbf{K} \in \mathbb{R}^{n \times n}$ for the whole collection. The similarities reveal how much a data point tells about all the other points of the set. A highly relevant measure is reflected by the notion of mutual information in information theory \cite{Cover2006}, which quantifies the amount of information obtained about one random variable by observing another one. Inspired by this notion, we require the similarity matrix to be symmetric, but relax the positivity constraint for each element to the \ac{psd} assumption on the matrix $\mathbf{K}$ for a more inclusive condition. Similarity measures widely used in statistics and machine learning, such as cosine similarity, Euclidean inner product, and all \ac{pd} kernel functions, satisfy these properties. 
Given this matrix, we aim at obtaining a representation matrix $\mathbf{R} \in \mathbb{R}^{n \times n}$ that encodes the representation power of each sample for describing others in the collection. Elements on a single row of this matrix, $\mathbf{r}^j$, correspond to the participation level of the $j$\textsuperscript{th} sample in representing the other data points. 
To achieve an optimal encoding, an optimization program is designed, in which we reward the representation power of the points capable of describing many others in the set, while penalizing the similarity among the chosen representatives. These criteria are realized through a linear and a quadratic term, respectively, as
\begin{equation}
\label{eq:obj_max} 
\max_{\mathbf{R}} \trace {2\mathbf{K}\mathbf{R} - \mathbf{R}^T\mathbf{K}\mathbf{R}}\:,
\end{equation} 
where we double the linear term to account for the symmetric forms obtained from the quadratic expansion. 
In order to reach a reduced subset of representatives, $\mathbf{R}$ is desired to be row-sparse. To this end, we transform our formulation to minimizing a cost function, coupled with a structured regularizer for row-sparsity:
\begingroup
\setlength\abovedisplayskip{3pt}
\setlength\belowdisplayskip{2pt}
\begin{equation}
	\label{eq:obj_quadratic} 
	\min_{\mathbf{R}} \frac{\lambda}{2} \trace {\mathbf{R}^T\mathbf{K}\mathbf{R}-2\mathbf{K}\mathbf{R}} + \| \mathbf{R}^T \|_{1,2}
\end{equation}
\endgroup
where $\lambda$ is an optimization parameter.
Framing the representation coefficients in a non-linear manner, along with the arbitrary \ac{psd} similarity matrix aims at capturing the intricate behavior of underlying manifolds. While modeling this complex relation, the designed program poses a well-defined convex problem, which makes the method independent of initialization.
The optimal solution $\mathbf{R^*}$ of this optimization problem is the desired encoding that carries the required structural information of the dataset underlying our ability to satisfy the three desired criteria for a representative set as described next.
\subsection{Locating Representative Samples}
\label{sec:representative}
The non-zero rows of $\mathbf{R}^*$ identify the points that best describe the whole collection, hence are better \textit{representatives}. Conversely, the ones whose corresponding rows are 
zero do not participate in representing the dataset efficiently, and hence, can be discarded from the subset. Additionally, each row reveals the influence level of the selected samples in representing the set; the ones corresponding to higher row-norms can be deemed as more influential ones, as it indicates more involvement of the corresponding sample in the reconstruction of the dataset. Their degree of involvement produces an influence-based ranking for the samples, which can be utilized to choose a limited number of samples, and also will be instrumental in choosing diverse representatives as described in the following.
\subsection{Diversity-based Pruning}
\label{sec:diversity}
Aside from the representativeness of the sampled subset, the efficacy of the sampling techniques are highly impacted by the diversity of information carried by chosen samples. Once more, inspired by information-theoretic metrics, we reason about this dissimilarity in terms of a metric called \textit{shared information distance}, which can be defined as $H(\mathbf{x}) + H(\mathbf{y}) -2 MI(\mathbf{x}|\mathbf{y})$, where $H(\cdot)$ and $MI(\mathbf{x}|\mathbf{y})$ denote the entropy of a variable and mutual information of $\mathbf{x}$ given $\mathbf{y}$, respectively \cite{Cover2006}. As the name implies, this metric gauges the amount of information lost and gained by moving from one point to another, and has been effectively used in comparison of different clusterings of a dataset \cite{Arabie1973,Meila2007}.
\subsection{Outlier Identification and Robust Sampling}
\label{sec:outliers}
To effectively cope with outlier contamination in practical applications, this section explicates how we identify various types of outlying data points present in the data as a final procedure to advance \textit{robustness} of the proposed sampling method to outliers.
To this end, we leverage the fact that outliers typically exhibit low total coherence to the rest of the data
\cite{Sedghi2018}.
Different notions of coherence or similarity have been utilized in the literature, however, intricate behavior of manifold data complicate the design of a distinguishing measure to effectively identify outlying data points. Inspired by \cite{Sedghi2019}, we exploit the richness of the obtained encoding to link the coherence of points to their ability to represent the collection.  
 In other words, each inlier is anticipated to be representable by a
few other inliers, while outliers are unlikely to follow this pattern, by being reconstructed mainly by themselves. 
Consequently, two groups of points are indicated by the non-zero rows of the optimal encoding: the influential inliers (which actively contribute to the reconstruction of others), and the outliers (that often cannot be represented by the rest of the samples rather than themselves). 
 Accordingly, we introduce a novel identification measure that can be cogently inferred from the sparsity of the rows of the optimal representation matrix. For a given sample $\mathbf{x}_j$ with a non-zero representation vector, we calculate its probability to be an outlier as
\begin{equation}
\ac{OP}(\mathbf{x}_j) \triangleq \frac{n-\frac{\|{\mathbf{r}^*}^j\|_1}{\|{\mathbf{r}^*}^j\|_\infty}}{n - 1}
\end{equation}
where ${\mathbf{r}^*}^j$ denotes the $j$\textsuperscript{th} row of the optimal representation matrix.
This quantity measures the involvement of a given sample in representing the other points of a dataset by its sparseness. Higher \ac{OP} values reflect the accumulation of the non-zero elements of the representation vector, suggesting higher probabilities for the sample to be an outlier. In contrast, inliers participating in the representation of many other inliers give rise to representation vectors with more evenly distributed elements, resulting in lower \ac{OP} values. This component can be integrated into the developed sampling method to reject the selected outliers out of sampled representatives, and also can be employed to detect outliers of a high-dimensional manifold-structured data autonomously.

%% file: Algorithm.tex
Generic solvers for convex problems such as \texttt{CVX} \cite{Grant2014,Grant2008} are known not to scale well with the problem size as they exhibit cubic or higher order complexities \cite{Elhamifar2014}. To alleviate this problem, we can readily use an ADMM or FISTA-based algorithm \cite{Boyd2011,Beck2009}, which considerable reduces the computational costs.  

\begin{algorithm}[t]
		\caption{Proposed Sampling Method from Manifolds}
		\label{alg:alg2}
		\begin{algorithmic}[1]
			\REQUIRE 
			kernel matrix $\mathbf{K}$, Optimization parameters.
			\ENSURE Sampled Representative Indices $\mathcal{I}$
			\STATE Obtain $\mathbf{R}^*$ via a convex solver as\\
			\begin{small}
			$\arg \min_{\mathbf{R}} \frac{\lambda}{2} \trace {\mathbf{R}^T\mathbf{K}\mathbf{R}-2\mathbf{K}\mathbf{R}} + \| \mathbf{R}^T \|_{1,2}$\\
			\end{small}
			\STATE $\mathcal{I} = \big \{j | \left\| \mathbf{r}^{*j} \right\|_2 \neq 0 \big\}$
			\STATE Keep only novel sample indices in $\mathcal{I}$ acc. to \sref{sec:diversity}
			\STATE Remove Outlier indices from $\mathcal{I}$ acc. to \sref{sec:outliers}
			\RETURN $\mathcal{I}$
		\end{algorithmic}
\end{algorithm}
\subsection{Scalable Implementation}
\label{sec:scalable}
In this section, we present a randomized scheme which yields a scalable implementation of the proposed method for large-scale data, which reduces the complexity to $\mathcal{O}(r^{1.373}\ceil{r/P})$, where $r << n$ is a parameter of this procedure. To this end, we initially sample a few $r$ random points of the original data uniformly, and apply our representative selection approach to the resulting subset. 
In order to compromise for possible information loss via random sampling, we devise an incremental refinement procedure to add back $\hat{r}$ maximally unrepresented data points to the random subset. More specifically, let $\mathbf{S}_c \in \mathbb{R}^{n \times r}$ denote the random sampling matrix, whose columns are zero except for one element of $1$ indicating the index of the sampled point. Then, $\mathbf{D}_c = \mathbf{D} \mathbf{S}_c \in \mathbb{R}^{m \times r}$ corresponds to the matrix of randomly sampled columns, and our objective function can be modified as \eqref{eq:objective_sampling}, where $\mathbf{K}_c = \mathbf{K}\mathbf{S}_c \in \mathbb{R}^{n \times r}$ and $\mathbf{K}_s = \mathbf{S}^T_c\mathbf{K}\mathbf{S}_c \in \mathbb{R}^{r \times r}$.
\begin{equation}
\label{eq:objective_sampling}
\frac{\lambda}{2} \trace {\mathbf{R}^T_c\mathbf{K}_s\mathbf{R}_c-2\mathbf{K}_c\mathbf{R}_c}+\| \mathbf{R}^T_c \|_{1,2}
\end{equation} 
After performing the optimization step on the randomly reduced matrix, an incremental refinement procedure is used to ensure enough columns are sampled by monitoring our objective function, which characterizes how far we are from optimally representing the dataset. More precisely, 
the misrepresentation cost for the $j$\textsuperscript{th} sample is 
\begin{equation}
\label{eq:rep_error}
e_j = k(\mathbf{d}_j, \mathbf{d}_j) -2k(\mathbf{d}_j, \mathbf{D}_c)\mathbf{r}^*_j + {\mathbf{r}^*_j}^T k(\mathbf{D}_c,\mathbf{D}_c) \mathbf{r}^*_j \:.
\end{equation}
Samples corresponding to high values of error have not been represented by the chosen ones satisfactorily, and hence, are added to the subset $\mathbf{D}_c$. This procedure is repeated for a few iterations to ensure high representability. The corresponding algorithm can be found in Algorithm \ref{alg:alg3}.
All the variables with the subscript $c$ correspond to their original variables with adapted size; \eg, $\mathbf{\Delta}_c \in \mathbb{R}^{n \times r}$.
\begin{algorithm} [t]    
		\caption{Proposed Scalable Randomized Algorithm for Sampling from Manifolds}          
		\label{alg:alg3}  
		\begin{algorithmic}[1]                    
			\REQUIRE Number of initial random samples $r << n$, number of added samples adaptively $\hat{r} << r$, number of iterations $I$.
			\ENSURE Sampled Representative Indices $\mathcal{I}$ 
			\STATE  \textbf{Initialization:}
			Set all optimization variables to zero
			\STATE Generate a zero matrix of size $n \times r$, except for one element of $1$ in each column, at random rows: $\mathbf{S}_c$.
			\FOR {$i = 1:I$}
			\STATE Form $\mathbf{K}_s = \mathbf{S}^T_c\mathbf{K}\mathbf{S}_c \in \mathbb{R}^{r \times r}$. 			
			\WHILE{\NOT $converged$} 
			\STATE 
			${\mathbf{\Delta}_c}_{(t+1)} = {(\lambda\mathbf{K}_s + \rho \mathbf{I}_c)}^{-1} (\lambda\mathbf{K}^T_c + \rho({\mathbf{R}_c}_{(t)} - {\mathbf{Q}_c}_{(t)}))$
			\STATE ${{\mathbf{r}}^i_c}_{(t+1)} = \left[1 - \frac{1/ \rho}{\left\| {({\mathbf{\delta}_c}_{(t+1)} + {\mathbf{q}_c}_{(t)})}^i \right\|_2}\right]_+ ({\mathbf{\delta}_c}_{(t+1)} + {\mathbf{q}_c}_{(t)})^i$
			\STATE ${\mathbf{Q}_c}_{(t+1)} = {\mathbf{Q}_c}_{(t)} + {\mathbf{\Delta}_c}_{(t+1)} - {\mathbf{R}_c}_{(t+1)}$
			\ENDWHILE
			\STATE $\mathbf{R}^*_c = \mathbf{R}^{k}_c$ \\
			\STATE Perform successive steps according to \S \ref{sec:diversity} - \ref{sec:outliers}. 
			\FOR {$j \in \mathbb{N}_n$}
			\STATE 
			$ e_j = k(\mathbf{d}_j, \mathbf{d}_j) -2k(\mathbf{d}_j, \mathbf{X}_c)\mathbf{r}^*_j + {\mathbf{r}^*_j}^T k(\mathbf{D}_c,\mathbf{D}_c) \mathbf{r}^*_j$ \COMMENT {Calculate the Misrepresentation Errors}
			\ENDFOR
			\STATE Find $\hat{r}$ highest errors and add the corresponding samples back to $\mathbf{D}_c$.
			\ENDFOR
			\RETURN $\mathcal{I}$
		\end{algorithmic}              	
\end{algorithm}

%% file: Analysis.tex
In this section, we present the main theorem of the paper, mainly built upon geometric functional analysis. For brevity, we denote the transformed images of a data point and the dataset by $\mathbf{y}_i = \phi(\mathbf{d}_i)$
and $\mathbf{Y} = \phi(\mathbf{D})$, respectively.

\begin{theorem}
	\label{thm:coverage}
	Define the notion of \textbf{coverage} of a set of points in $\mathcal{S}$ by a subset $\bar{\mathcal{S}}\subseteq \mathcal{S}$ as $\sum_{\mathbf{s}_j \in \mathcal{S}} \operatorname{len}(\vec{\mathbf{s}}_j \cap \mathcal{P}(\bar{\mathcal{S}})) = \sum_{\mathbf{s}_j \in \mathcal{S}} \|\hat{\mathbf{s}}_j\|_{\mathcal{H}}$, where $\hat{\mathbf{s}}_j$ is the intersection point of the sample $\mathbf{s}_j$ and the convex hull of the collection $\mathcal{P}(\mathcal{S})$, and $\operatorname{len}(.)$ is the length of the corresponding line segment. Minimizing our objective function $F_{1,\infty} (\mathbf{K}, \mathbf{R})$ is equivalent to maximizing the \textit{coverage} of the whole collection $\mathcal{Y}$ by our sampled representatives $\bar{\mathcal{Y}}$. 
\end{theorem}
Geometrically, our objective function measures how well the collection of data is \textit{covered} by the sampled representatives. In the supplementary material throughout the proof of the theorem, this is illustrated in a schematic figure, where the sampled representative set is $\bar{\mathcal{Y}} = \{ \mathbf{y}_1,\ldots, \mathbf{y}_4\}$ shown in blue points, and $\mathbf{y}_j$ is an arbitrary point of the collection. Our proposed method maximizes the length of the line segments shown in red for each point of the data collection, which intuitively translates to maximizing the data points under \textit{coverage} of the sampled representatives, for the whole dataset. Poorer representations result in higher values of the objective function, which implies that many of the data points cannot be covered by the chosen samples.
\begin{theorem}
	\label{thm:simplex}
	Consider a simplex whose vertices are inside the transformed dataset. Minimizing the objective function of \thmref{thm:coverage} correspond to finding the extreme points of the union of all such simplices. As such, the non-zero rows of the optimal solution are at one-to-one correspondence to these extreme points. 
\end{theorem} 

Additional characterizations of the identified representatives, as well as the proofs of the main and intermediate results are provided in the supplementary material. We also present insightful interpretations of the aforementioned analysis which further clarifies our methodology. 

%% file: Experiments.tex
In what follows, we present illustrating experiments to assess the key characteristics of our method. The experiments are carried out on both synthetic and real benchmark manifold datasets as explained in each section. We also conduct extensive comparisons between our method and related state-of-the-art algorithms designed to handle manifold data  All the parameter settings were chosen as suggested by the authors. 
Also, for all random initializations/data generation procedures, we report the average results over $50$ runs. 
The data is randomly split to training ($70\% $), validation ($10\% $), and testing sets ($20\% $).
The best parameter is chosen over the validation set.
The first set of experiments aim to showcase the qualitative performance of our method for sampling out of clustered data. In the second set of experiments, various sampling algorithms are applied to different datasets, and then we conduct numerical comparisons among different approaches for several classification tasks using the obtained representatives for both synthetic and real data. 
Furthermore, robustness of our sampling method to different types of outliers including 
random, repetitive, and structured outliers is examined in Sec. \ref{sec:sampling_robustness}. The efficiency of the developed optimization algorithms are analyzed through comparison of running times with a generic convex solver as well as other methods. Lastly, additional experiments are included in Sec. \ref{sec:outlier_exp}, comparing the performance of outlier detection technique with related methods. 

\subsection{Clustering Applications}
To illustrate how our method samples representatives in a fair manner from clustered datasets, we showcase two real-world examples of Face sampling. We consider the Extended Yale Database \cite{Georghiades2001} to sample from a randomly chosen subset of $5$ persons.
\begin{figure}[t]
	\centering
	\vspace{-1.9cm}
	\includegraphics[width=0.45\textwidth]{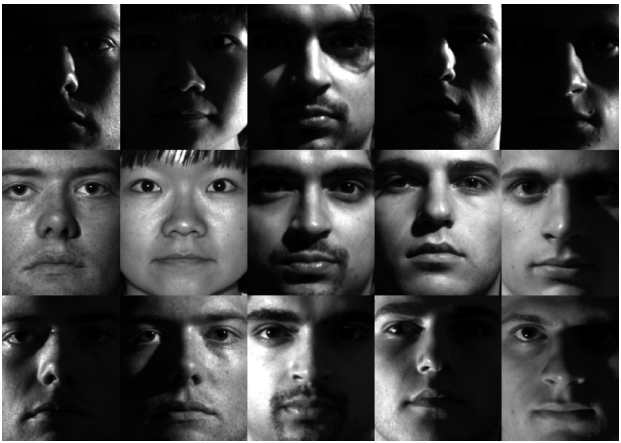}%
	\vspace{-2.3cm}
	\caption{Sampling from $5$ random subjects of Extended Yale. The representatives are chosen adequately form each cluster. Each row follows a specific pattern of illumination; right, straight, and left from top to bottom.}
	\label{fig:face_cluster}
\end{figure}
The result in \fref{fig:face_cluster} reveals that we sample an almost equal number of samples out of each cluster.  
Sampled representatives induce the interesting observation that, each row of the figure seems to exhibit a particular pattern for each subject. More specifically, the middle row corresponds to the face images of these subjects under normal lighting, while the top and bottom row correspond to the images with illuminations from right, and left hand-side, respectively. One can interpret that our method attempts to capture different aspects of the face images by focusing on different angles to gain a global perception of the face characteristics of each subject. As a result, our method samples representatives with different emphasis of right, straight, and left side of the faces, from top to bottom.

\subsection{Robustness to Simple and Structured Outliers}
\label{sec:sampling_robustness}
\begin{figure}[t]
	\centering
	\hspace{-0.1cm}
	\subfloat[]{\label{fig:sphere_outlier}\includegraphics[width=0.25\textwidth]{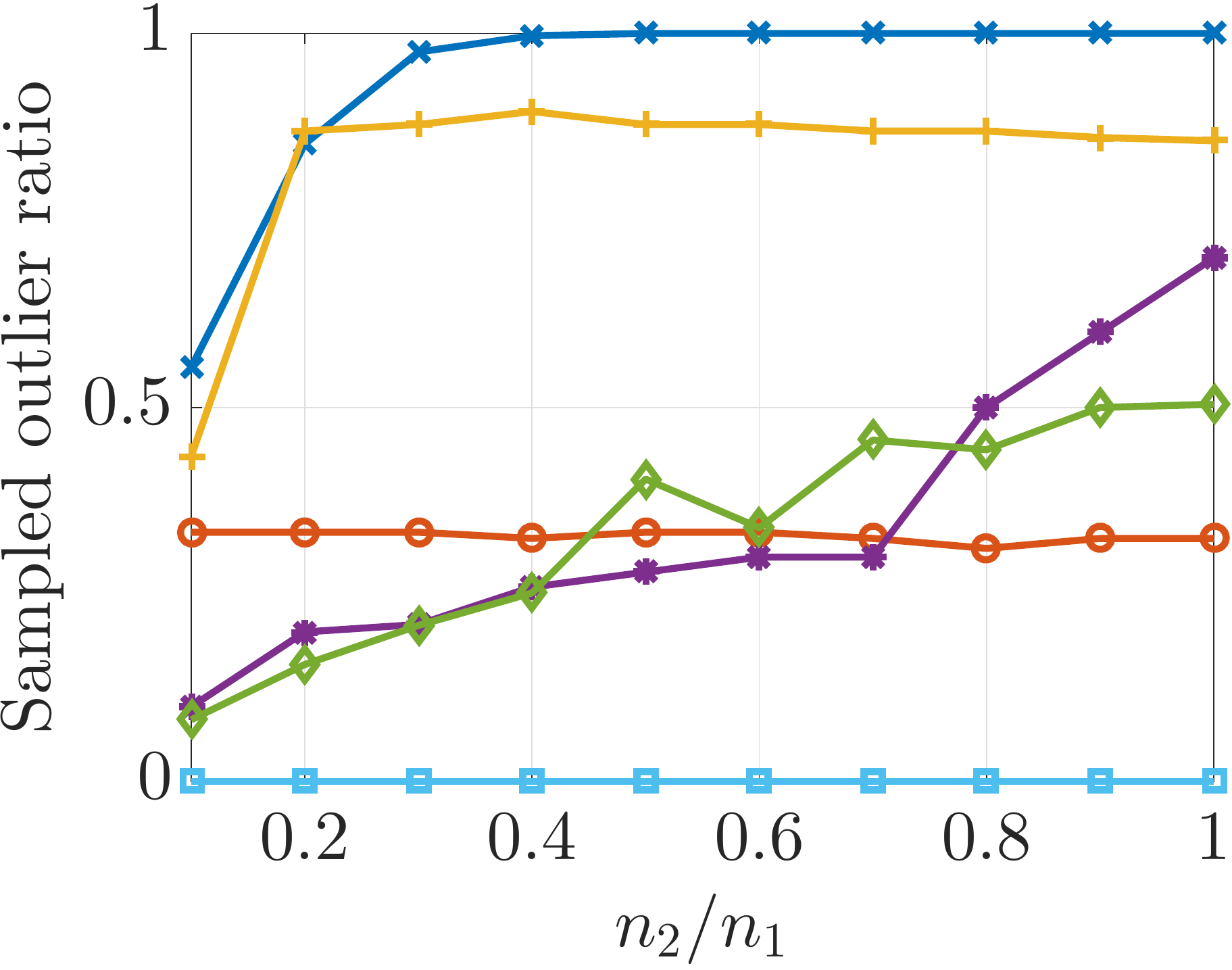}}
	\subfloat[]{\label{fig:knot_outlier}\includegraphics[width=0.24\textwidth]{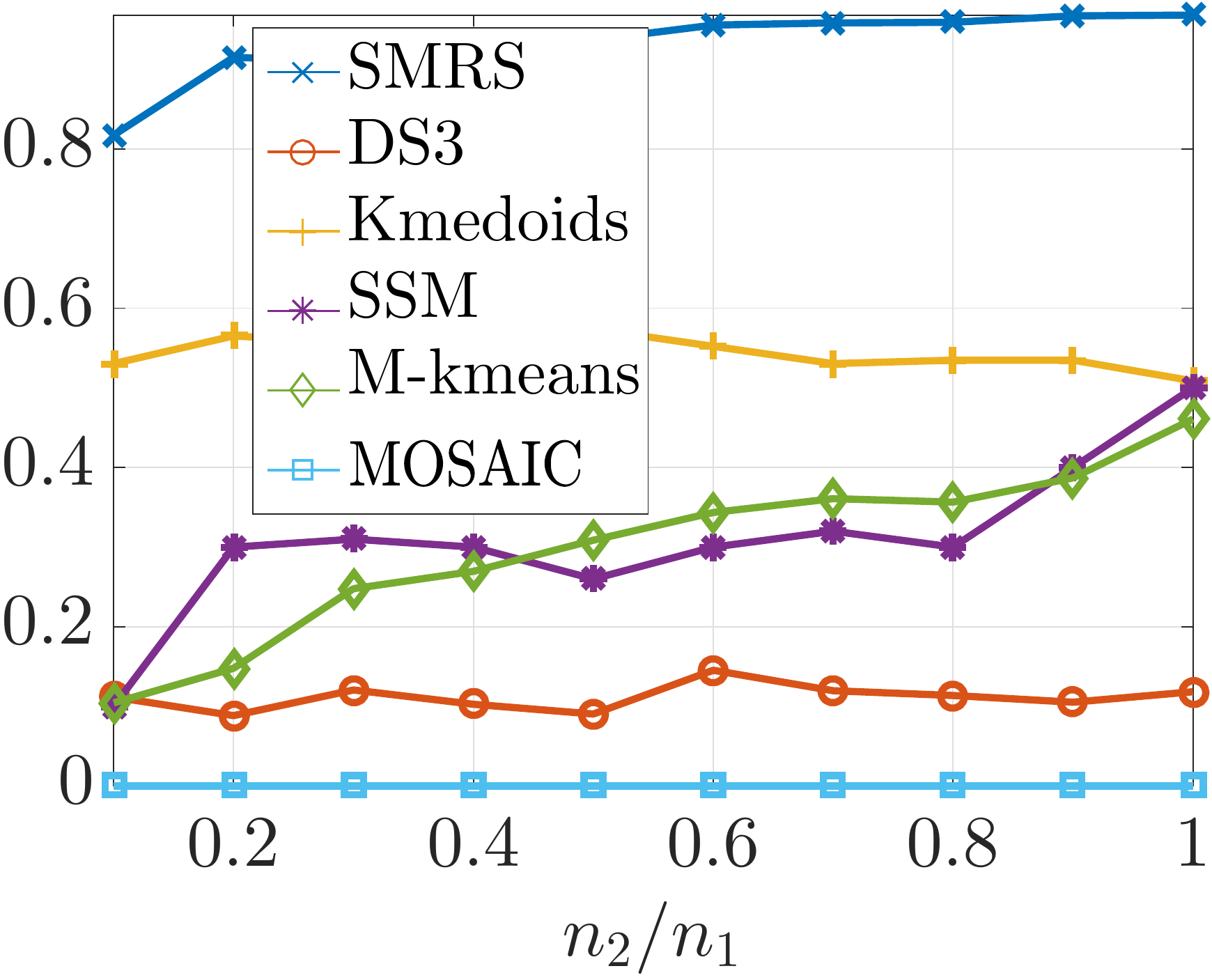}}\\
	\vspace{-0.3cm}
	\subfloat[]{\label{fig:SwissRoll_outlier}\includegraphics[width=0.25\textwidth]{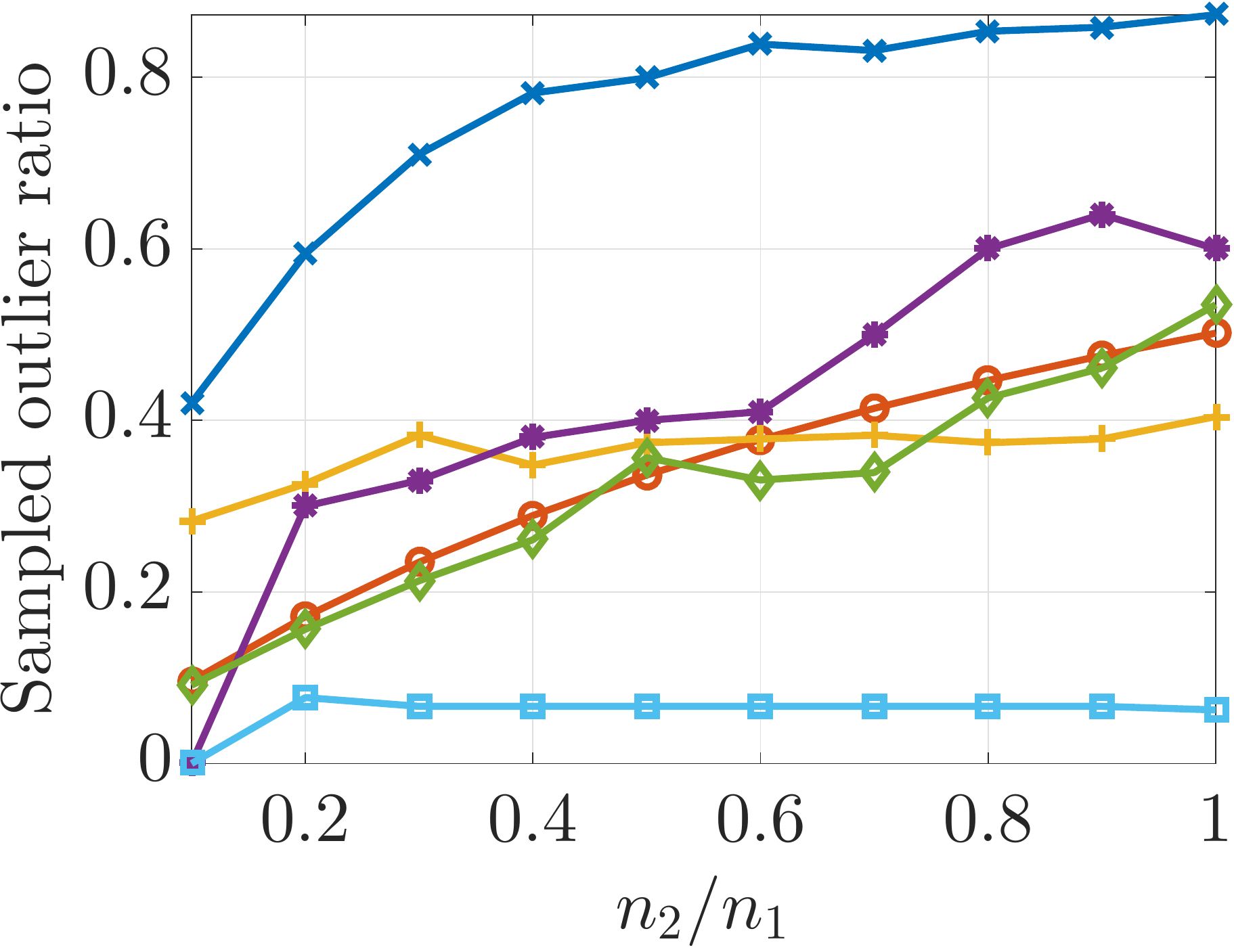}}
	\subfloat[]{\label{fig:face_outlier}\includegraphics[width=0.24\textwidth]{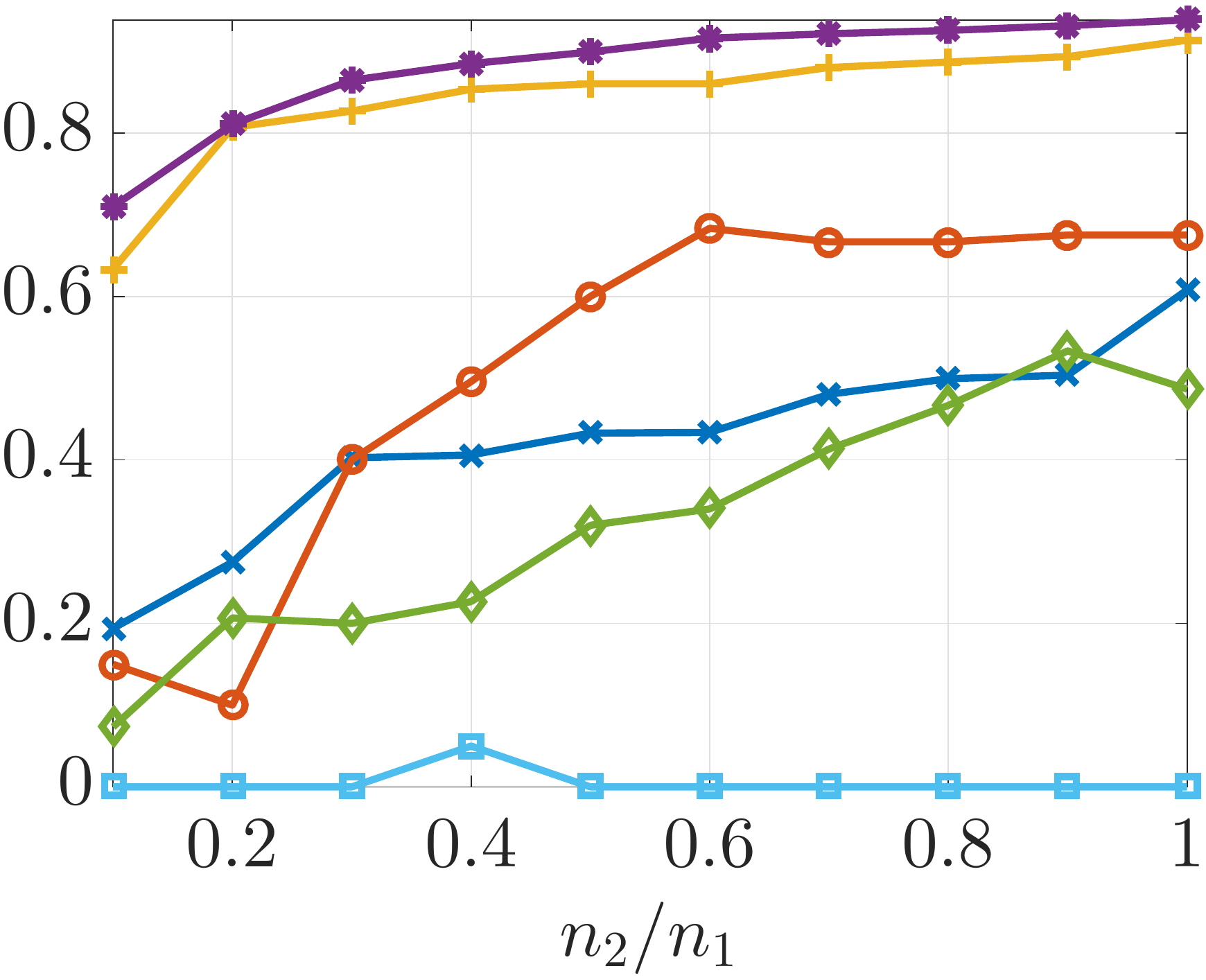}}
	\caption{Robustness to unstructured (first row)/structured (second row) outliers. Random outliers for \protect\subref{fig:sphere_outlier} Sphere, \protect\subref{fig:knot_outlier} Trefoil Knots datasets. Repetitive outliers for \protect\subref{fig:SwissRoll_outlier} SwissRoll, and random image outliers \protect\subref{fig:face_outlier} Frey Face dataset.
		The proposed method \ac{name} substantially outperforms the other approaches under various outlier types.}
	\label{fig:outlier}
\end{figure}

To evaluate the robustness of our sampling method to outliers, we perform the repersentative selection procedure on contaminated data. In presence of random outlying points, an outlier matrix of $n_2$ uniformly sampled points from $\mathbb{R}^{m}$ is concatenated to the datasets. The ratio of outliers to inliers is incremented from $0.1$ to $1$ in \fref{fig:outlier}.  
We first consider simple outliers, consisting of uniformly random points in the ambient space of the Sphere and Trefoil Knots datasets. These two datasets are both synthesized such that contain spherical, and knot manifolds embedded n a high-dimensional space. Furthermore, in order to evaluate the method's performance in a more challenging setting, we experiment how it tolerates structured outliers for both artificial and real data.
We design two different types of structured outliers for different datasets. First, to impose a linearly dependent structure to the outlier matrix, we set $0.1$ of the $n_2$ uniformly random outliers to be repetitive points, to be concatenated to the SwissRoll dataset. This is another artificially generated data in which the benchmark swissRoll manifold is embedded. This dependency makes the outliers representable by each other, but we still anticipate them to bear weak resemblance to the inliers in terms of their participation in the representation of the whole dataset. Second, for real-life data, we consider the Frey Face datatset, which is a publicly available $560$-dimensional dataset (see https://cs.nyu.edu/roweis/data.html), which consists of $2000$ grayscale face images of a person taken from the frames of a short video. Images are $20\times28$, representing diverse facial expressions taken from different angles.  For this experiment we choose $n_2$ random natural images downloaded from the Internet, convert them to grayscale, and resize them to match Frey Face images to be concatenated to them. This outlier matrix is known to carry structural information due to natural image statistics \cite{Hyvrinen2009}.
\ac{name} is remarkably robust to various types of outliers, and substantially outperforms the other approaches as the number of outliers increase.
\subsection{Scalable Algorithms' Efficiency}
\label{sec:time}
\begin{figure}[t]
 	\centering
 	\includegraphics[width=0.35\textwidth]{
 		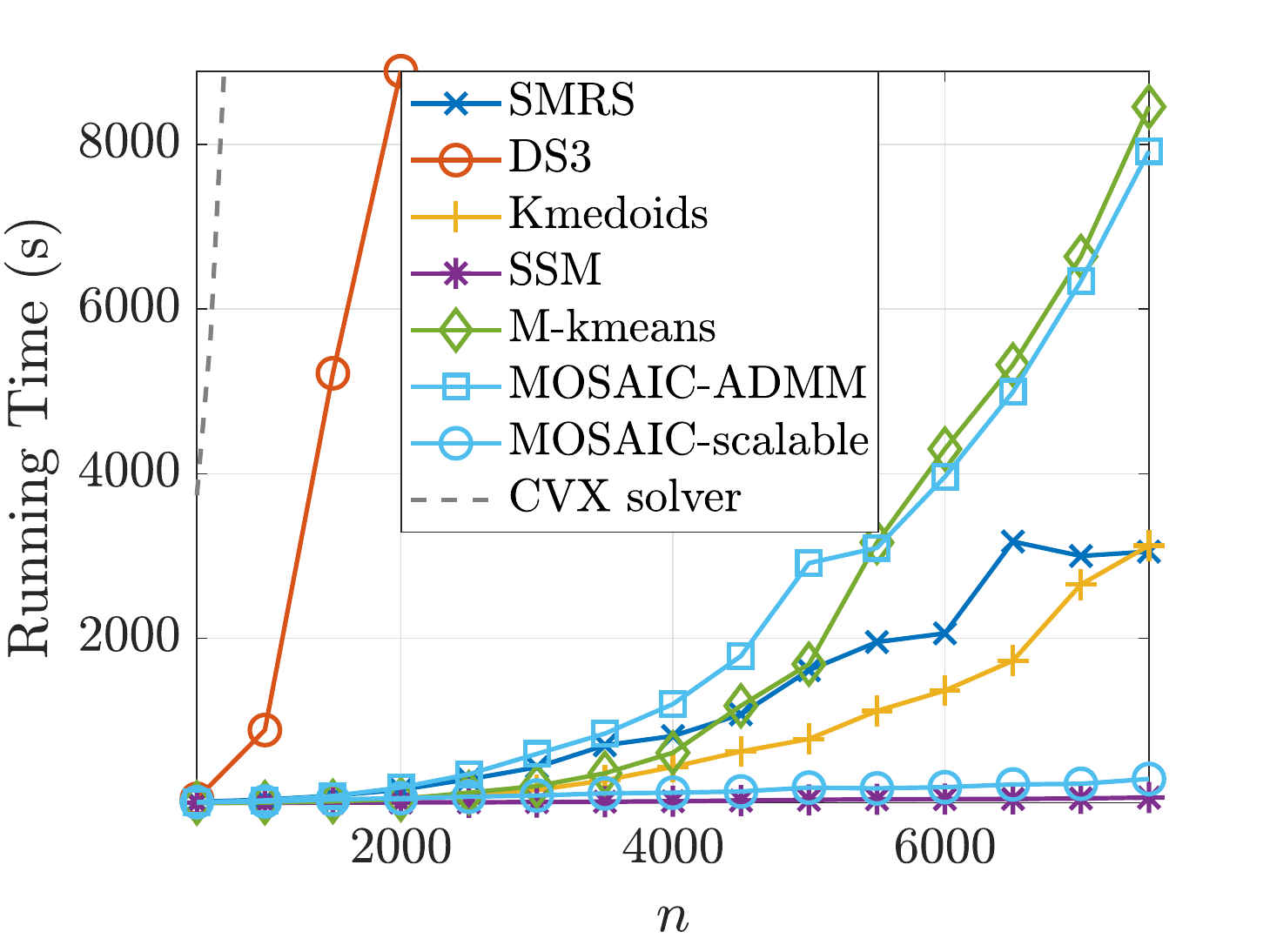}
 	\caption{Running time comparison among all sampling methods, CVX solver, and the developed \ac{ADMM}-based and scalable algorithm.
 	}
 	\label{fig:time_MSRB}
 \end{figure}

Although generic convex solvers may be powerful, they do not scale well with the problem size.
To boost the efficiency of our method even further, a randomized scalable scheme was proposed in \sref{sec:scalable}. Gained improvements from our developed algorithms are investigated by monitoring their running time versus other methods for large-scale subsets of \textit{MSRA-B} dataset, shown in \fref{fig:time_MSRB} \cite{Wang2017}. This is a widely used object detection/segmentation dataset, which provides salient object annotations for $5000$ RGB images of size $300\times400$. The gray dashed line indicates the running time of the \texttt{Sedumi} solver of the \texttt{CVX} Matlab toolkit for solving our optimization, which is significantly slower than the ADMM algorithm. The better efficiency of our scalable algorithm, designated by \ac{name}-scalable, is reflected by its substantial saving in computational time. Note that as can been seen from our classification results, our method does not compromise much accuracy in order to earn this speedup.  
Our algorithm runs extremely faster than the others, except for SSM, which exhibits similar running times. 
However, our approach is more powerful in that it outperforms this method with regards to other performance metrics (\eg, classification accuracy).  
The experiments are conducted on an 
X$64$-based system with 
$2.4$ GHz CPU and 
$32$ GB memory.
\subsection{Stand-alone Outlier Identification}
\label{sec:outlier_exp}
Our proposed framework yields a decomposition of the data into points lying on a union of manifolds (inliers) and the ones not adhering to these manifolds (outliers), giving rise to a stand-alone technique for detecting outlying points. This section provides additional experiments presenting an application of the proposed technique, and scrutinizing its behavior in presence of various outlier types. For comparison, we have used state-of-the-art outlier detection methods for nonlinear data including RML \cite{Du2013}, FDR \cite{Ye2015}, and $L_1$-KPCA \cite{Kim2017}, and their parameters are set as suggested by the authors. 
\subsubsection{Salient Object Detection}
\label{sec:salient}
\begin{figure}[b]
	\includegraphics[width=0.5\textwidth]{
		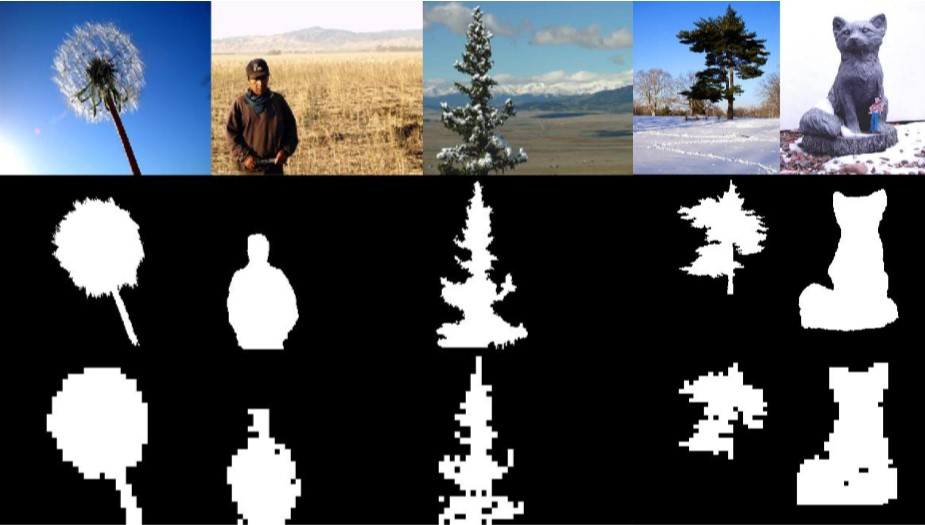}
	\caption{Application of the proposed method to Salient Object Detection in MSRA-B dataset. Second row: ground-truth, third row: our result.}
	\label{fig:salient}
\end{figure}
 The aforementioned two-way decomposition of data can be utilized in a variety of applications of this nature. We showcase the salient object detection application as an important computer vision task for images from the MSRA-B dataset \cite{Koch1987}. To do so, each data point is taken to be the concatenation of small patches of an image. One can expect these patches to be highly similar to each other, hence forming an internal structure for the outliers. The samples identified as outliers by our method reveal the saliency map of the images. Some examples of the generated maps are displayed in \fref{fig:salient}, where $8 \times 8$ non-overlapping patches are extracted. Although no vision-specific pre-processing has been done on the data, our results properly match the visually salient regions of the images. 
 The global understanding of the underlying patterns advances such an effective identification technique.
\subsubsection{Structured Outlier Detection}  
\label{sec:structured}
\begin{figure}[t]
	\centering
	\includegraphics[width=0.35\textwidth]{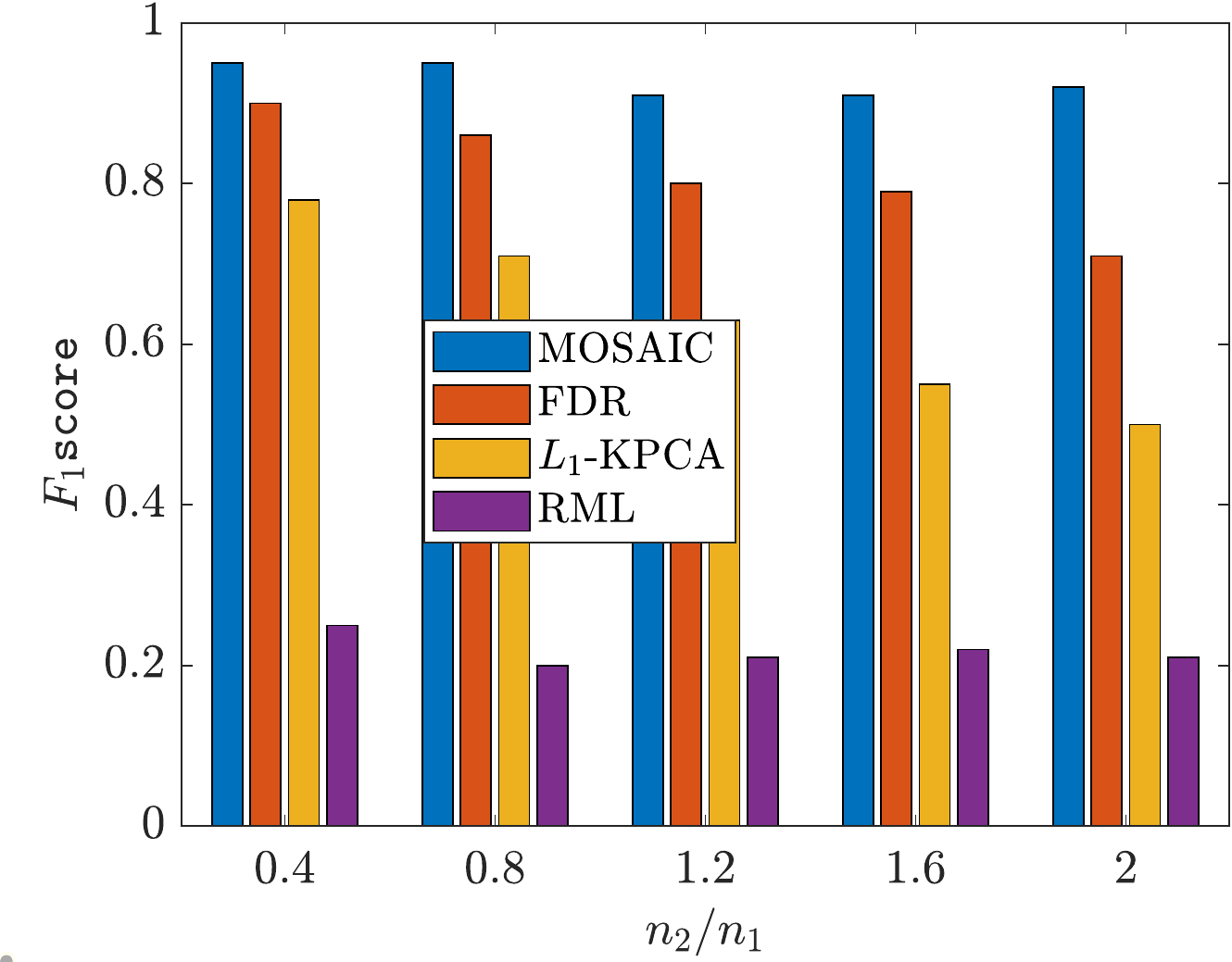}
	\caption{Performance of different algorithms in presence of structured natural image outliers, for different ratios of $n_2/n_1$ (Frey Face dataset).}
	\label{fig:face_natural_outlier}
\end{figure} 
\begin{table}[b]
	\caption{Performance Comparison ($F_1 \mathtt{score}$) of different algorithms for SwissRoll dataset, with mixed (simple and linearly dependent) outliers. 
	}
	\label{Table:comparison}	
	\centering	
	\normalsize
	\resizebox{0.5 \textwidth}{!}{
		\begin{tabular}{ |c |c  | c | c | c | c |}
			\hline
			Method \textbackslash $n_2/n_1$ & $1$ & $2$ & $3$ & $4$ & $5$ \\
			\small \textbf{\ac{name}} &$\mathbf{0.930}$ & $\mathbf{0.889}$ & $\mathbf{0.851}$& $\mathbf{0.816}$ & $\mathbf{0.796}$\\
			\hline
			\small FDR & $0.889$&$0.779$&$0.566$&$0.449$&$0.380$\\ 
			\hline
			\small $L_1$-KPCA & $0.907$ & $0.741$ & $0.601$ & $0.509$ & $0.309$\\ 
			\hline
			\small RML & $0.828$ & $0.639$ & $0.533$ & $0.454$ & $0.347$\\ 
			\hline
		\end{tabular}	
	}
\end{table}
We compare the performance of the proposed scheme with state-of-the-art outlier detection methods. 
Similar to \sref{sec:sampling_robustness}, we experiment with random plus linearly dependent outliers as well as statistically correlated points. 
\tref{Table:comparison} shows the identification results for the random plus linearly dependent case for the SwissRoll dataset. The ratio of outliers to inliers start from $1$ and go up to $5$ by increments of $1$. Although showing acceptable performance at first, all other methods lose track of the outliers when their number increases, leading to degraded performance. 
\fref{fig:face_natural_outlier} reflects how our method steadily outperforms all other algorithms in the case of the real-life example of Frey Face database contaminated by natural images. Among others, \ac{FDR} has the second place in robustness to this type of structured outliers, but because of the sparsity assumption, its performance deteriorates when the number of outliers grow. 

%% file: Conclusion.tex
We proposed \ac{name}, a powerful, scalable and robust approach to representative selection, tailored for non-linear manifold data. By considering global manifold structures, \ac{name} demonstrates a preeminent behavior through selection of descriptive and diverse representatives, while rejecting disruptive information. The outlier detection procedure is remarkably robust to challenging outlier settings such as large number of simple outliers and structured ones. Also, a randomized scalable implementation is proposed which offers substantial speedups compared to other methods. 
Mathematical analysis illustrates the geometrical characteristics of the solution, shedding light on the superior performance of the method in various experiments on both synthetic and real datasets.